\documentclass[10pt,twocolumn,letterpaper]{article}

\usepackage{iccv}              

\usepackage[dvipsnames]{xcolor}
\usepackage{colortbl}

\newcommand{\ours}[0]{{\color{black}FLIPNET}\xspace}
\newcommand{\jattn}[0]{{\color{black}BoostHub}\xspace}
\newcommand{\RM}[0]{{\color{black}restoration mode}\xspace}
\newcommand{\DM}[0]{{\color{black}degradation mode}\xspace}

\usepackage{times}
\usepackage{graphicx}
\usepackage{amsmath}
\usepackage{amssymb}

\usepackage{subcaption}
\usepackage{multirow}
\usepackage{bm}
\usepackage{ifthen}

\usepackage{makecell}
\usepackage{mathtools}

\usepackage{array}
\newcolumntype{C}[1]{>{\centering\arraybackslash}p{#1}}

\definecolor{realpink}{RGB}{250,127,122}
\definecolor{ourblue}{RGB}{91,155,213}
\definecolor{lightpurple}{HTML}{d9d6f6}
\definecolor{lightpink}{HTML}{fde0e9}

\newcommand{\bgcgrey}[1]{{\cellcolor[HTML]{F2F2F2}{#1}}} 

\newcommand{\bred}[1]{\textbf{\color[HTML]{CB0000}{#1}}}

\newcommand{\tightcolorbox}[2]{\begingroup\fboxsep=1pt\colorbox{#1}{#2}\endgroup}
\newcommand{\onres}[1]{\tightcolorbox{lightpurple}{#1}}
\newcommand{\onour}[1]{\tightcolorbox{lightpink}{#1}}

\DeclareMathOperator{\N}{\mathcal{N}}

\definecolor{iccvblue}{rgb}{0.21,0.49,0.74}
\usepackage[pagebackref,breaklinks,colorlinks,allcolors=iccvblue]{hyperref}
\usepackage{float}
\usepackage{ulem}
\usepackage{arydshln}
\usepackage{todonotes}
\usepackage{xcolor}

\def\paperID{3541} 
\def\confName{ICCV}
\def\confYear{2025}

\title{Unlocking the Potential of Diffusion Priors in Blind Face Restoration}

\author{
  Yunqi Miao\textsuperscript{1}\quad
  Zhiyu Qu\textsuperscript{3}\quad
  Mingqi Gao\textsuperscript{2}\quad
  Changrui Chen\textsuperscript{1}\quad \\
  Jifei Song\textsuperscript{3}\quad
  Jungong Han\textsuperscript{2}\quad
  Jiankang Deng\textsuperscript{4} \\
 \textsuperscript{1}University of Warwick \quad
  \textsuperscript{2}University of Sheffield \quad
  \textsuperscript{3}University of Surrey \quad
  \textsuperscript{4}Imperial College London \\
  {\tt\scriptsize $\{$yunqimiao709, im.mingqi, jungonghan77, jiankangdeng$\}$@gmail.com, changrui.chen@warwick.ac.uk, z.qu@surrey.ac.uk}
}

\begin{document}
\maketitle
\begin{abstract}

Although diffusion prior is rising as a powerful solution for blind face restoration (BFR), the inherent gap between the vanilla diffusion model and BFR settings hinders its seamless adaptation. 
The gap mainly stems from the discrepancy between 1) high-quality (HQ) and low-quality (LQ) images and 2) synthesized and real-world images.
The vanilla diffusion model is trained on images with no or less degradations, whereas BFR handles moderately to severely degraded images.
Additionally, LQ images used for training are synthesized by a naive degradation model with limited degradation patterns, which fails to simulate complex and unknown degradations in real-world scenarios.
In this work, we use a unified network \ours that switches between two modes to resolve specific gaps.
In \RM, the model gradually integrates BFR-oriented features and face embeddings from LQ images to achieve authentic and faithful face restoration.
In \DM, the model synthesizes real-world like degraded images based on the knowledge learned from real-world degradation datasets.
Extensive evaluations on benchmark datasets show that our model 1) outperforms previous diffusion prior based BFR methods in terms of authenticity and fidelity, and 2) outperforms the naive degradation model in modeling the real-world degradations. 

\begin{figure}
    \centering
    \includegraphics[width=\linewidth]{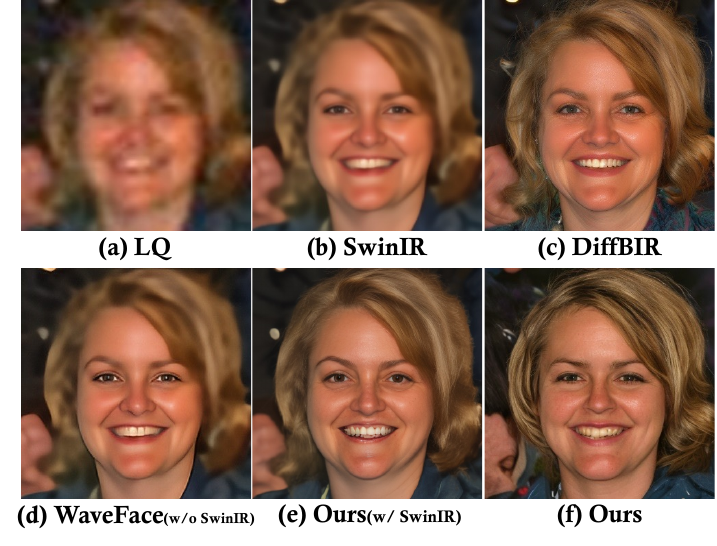}
    \vspace{-7mm}
    \caption{Restoration results of WIDER~\cite{yang2016wider}. Pre-processing module (\ie SwinIR) based methods (\ie DiffBIR) fail to render fine-grained details. Methods with condition-concatenated inputs (\ie WaveFace) unable to handle unknown and complex degradation in real-world scenarios, leading to over-smoothed results.}
    \label{fig:comp}
    \vspace{-7mm}
\end{figure}
\end{abstract}
    
\vspace{-4mm}
\section{Introduction}
\label{sec:intro}
Blind face restoration (BFR) aims to recover high-quality (HQ) facial images from various degradations.
The ill-posed nature of this task stems from the complexity and uncertainty of degradations in real-world scenarios.
Versatile facial priors have been used to help BFR, such as facial landmarks~\cite{chen2018fsrnet}, referential HQ images or features~\cite{wang2023restoreformer++,tsai2024daefr}, and the pretrained StyleGAN~\cite{yang2021gpen, wang2021gfpgan}.
The success of diffusion models (DMs)~\cite{dhariwal2021diffusion, rombach2022ldm} in generation tasks has recently excited the restoration field, as their ability to generate HQ content from noise aligns perfectly with the restoration objective~\cite{wang2023dr2,lin2023diffbir,yang2023pgdiff,miao2024waveface}.
Despite its powerful generative ability, the inherent gap between the vanilla diffusion process and the BFR setting hinders its seamless adaptation.
The gap arises mainly from the distribution mismatch between: 1) HQ images and low-quality (LQ) images and 2) synthesized images and real-world images.

The vanilla DMs are trained on natural images with minor or no degradation, making them unfit the BFR task, which primarily targets moderately to severely degraded face images.
To bridge the gap, previous attempts~\cite{lin2023diffbir, oh2025pmrf} brutally remove the severe degradations via a pre-processing module~\cite{liang2021swinir}, followed by a DM to render facial details.
However, the pre-processing module also removes discriminative details such as wrinkle and freckle (\cref{fig:comp}(b)), thus compromising the fidelity of restoration (\cref{fig:comp}(c)).
Some approaches~\cite{miao2024waveface} bypass the pre-processing module by concatenating the degraded images, acting as the condition, to HQ inputs.
Although the first gap is mitigated by gradually introducing LQ features throughout denoising, they fail to deliver adequate results on real-world data (\cref{fig:comp}(d)) as the degradation model used for training set synthesizing fails to simulate real-world degradations.
A pioneering work~\cite{wang2021realesrgan} applies the naive degradation model repeatedly to HQ images to create LQ counterparts.
Although effective, this work generates less diverse data due to its reliance on limited degradation patterns, which is incompatible with ``blind'' face restoration.

To address these problems, this work uses only a large-scale Text-to-Image (T2I) model with few trainable parameters, eliminating the reliance on pre-processing modules in both training and inference.
Our goal is to steer the pretrained T2I model towards authentic and faithful face restoration while preserving its strong image prior by properly integrating contributive features from LQ images.
Despite offering valuable cues such as color and structure for faithful face restoration, LQ images also carry the degradation that affects authenticity.
To strike a balance, we propose \jattn to selectively integrate BFR-oriented features from LQ inputs while discarding irrelevant ones.
Meanwhile, we identify the misalignment between face embeddings used in personalized T2I generation~\cite{papantoniou2024arc2face, peng2024portraitbooth} and those needed for BFR, which inspires us to explore BFR-oriented face embeddings.
Apart from face restoration, we unlock the potential of diffusion prior in modeling real-world degradations.
Being trained with HQ and degraded image pairs collected from real-world scenarios, the T2I model learns the distribution of real-world degradations, which is used to synthesize degraded images.

In this work, we unify the above solutions into a single model, \ie \ours, which takes HQ and LQ image pairs as input.
By simply flipping the inputs, the model switches between \RM and \DM, achieving authentic and faithful face restoration and real-world degradation synthesis, respectively.
The contributions of this work can be summarized as follows:
\begin{itemize}
    \item A two-pronged solution, \ie \ours, is proposed to bridge two inherent gaps that hinder the seamless adaptation of diffusion prior in BFR by simply switching between restoration mode and degradation mode.
    \item When switching to \RM, \ours achieves authentic and faithful face restoration by integrating BFR-oriented features and face embeddings from LQ images.
    \item When switching to \DM, \ours learns from degraded images in real-world scenarios and, in turn, synthesizes such images with acquired knowledge.
    \item Comprehensive experiments demonstrate the superiority of \ours in integrating diffusion prior to BFR.
\end{itemize}

\section{Related works}
\noindent{\bf Blind face restoration (BFR).}
BFR works are generally based on three priors: 
1) \textbf{Geometric priors}, including facial landmarks~\cite{chen2018fsrnet,deng2019menpo} and facial parsing maps~\cite{chen2021progressive,zheng2022decoupled}, are used to provide explicit facial structure information.
They often yield inferior restoration results due to the imprecise prediction on degraded images.
2) \textbf{Reference priors} based methods either leverage high-quality (HQ) reference images to provide identity information~\cite{varanka2024pfstorer} or construct a dictionary that restores HQ facial features~\cite{wang2023restoreformer++, tsai2024daefr}.
But the restoration performance is restricted by the authenticity of reference images and the size of the codebook.
3) \textbf{Generative priors} encapsulated in StyleGAN2~\cite{karras2020stylegan2} are used to provide rich facial details for restoration. GFP-GAN~\cite{wang2021gfpgan} and GPEN~\cite{yang2021gpen} restore degraded faces by learning the mapping from LQ input to the latent space where GAN prior network can reproduce the desired HQ face images.

\noindent{\bf Diffusion prior in BFR.}
Inspired by the ability of diffusion models (DMs)~\cite{song2020denoising,rombach2022ldm} in emerging HQ content from noise, diffusion prior has become a favorable solution for BFR.
However, DMs are typically trained on images with less or no degradation while BFR deals with moderately to severely degraded images, which makes its seamless adaptation challenging.
Therefore, some works break BFR down to two stages: degradation removal and details generation.
DR2~\cite{wang2023dr2} employs the diffusion prior at first stage to obtain degradation-invariant images, where HQ counterparts are obtained by the follow-up face restoration model.
DiffBIR~\cite{lin2023diffbir} and PMRF~\cite{oh2025pmrf} first coarsely remove severe degradations with a pretrained pre-processing module~\cite{liang2021swinir}.
Given the obtained intermediate results, DiffBIR renders realistic facial details via a follow-up T2I model.
Whilst PMRF takes such images as the initial state of the distribution transportation, where a rectified flow model is optimized to model the path to HQ counterparts.
Instead of adopting such a two-stage scheme, PGDiff~\cite{yang2023pgdiff} uses a pretrained BFR model~\cite{wang2021realesrgan} to provide instructive semantics. Throughout the denoising process, the model is trained to minimize the discrepancy between intermediate outputs and target properties.
WaveFace~\cite{miao2024waveface} uses diffusion prior to restore the low-frequency component exclusively, where low-frequency subbands of LQ images are concatenated to noise and serve as condition for generation.
Although exciting restoration results have been achieved, the pre-processing module will also smooth out facial details, thereby compromising the fidelity.
Additionally, the reliance on other face restoration models imposes limitations on the potential of the proposed model. 
How to effectively unleash the potential T2I model in BFR task has NOT been fully exploited.

\noindent{\bf Real-world degradation modeling.}
Apart from the gap between HQ and LQ images, diffusion prior BFR methods general deliver inferior performance on real-world images as the degradation model used to synthesize training set fail to simulate that in real-world scenarios.
Classical degradation model~\cite{li2018learning} includes four types of degradations: blur, noise, resize, and JPEG compression.
The simple combination fails to handle real-world cases where images typically spread several times over different digital devices.
Thus, Real-ESRGAN~\cite{wang2021realesrgan} proposes a ``high-order'' degradation model, where classical degradations with different hyper-parameters are applied repeatedly on HQ images to synthesize the degraded counterparts.
While this may serve as a stopgap, the synthesis remains constrained by the limited degradation patterns.
How to effectively simulate complex and blind degradations remains a formidable challenge.

In this work, we address above challenges with a unified model that is solely based on a large-scale T2I model with few trainable parameters.
The model provides a two-pronged solution to both authentic and faithful face restoration and real-world degradation modeling.

\begin{figure*}[!htbp]
    \centering
    \includegraphics[width=\linewidth]{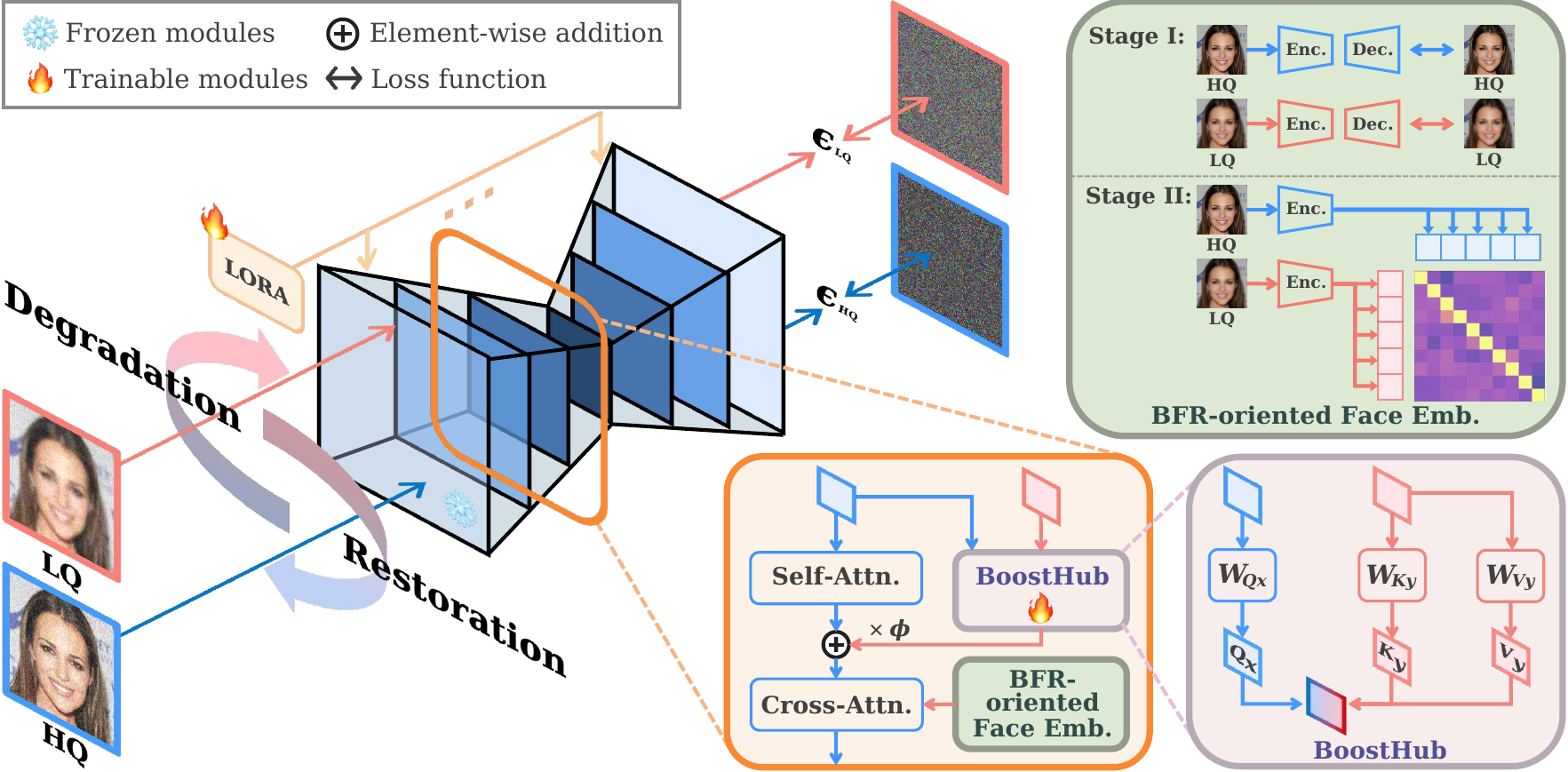} 
    \caption{\textbf{\ours pipeline.} \ours takes noisy high-quality (HQ) and low-quality (LQ) image pairs as input. It can switch between \RM and \DM. Taking \RM as an example, HQ images are used as input while LQ images are condition, where a \jattn is placed in parallel to each self-attention layer to selectively integrate the BFR-oriented LQ features for denoising. Additionally, LoRA weights are plugged to all self-attention and cross-attention layers to adapt the base model to face domain. By simply flipping the image-condition order, it switches to \DM to synthesize degraded data with real-world degradations.}
    \label{fig:frame}
    \vspace{-4mm}
\end{figure*}

\section{Preliminary}
\noindent\textbf{Stable Diffusion} is a large-scale text-to-image (T2I) latent diffusion model, which performs both diffusion and denoising processes in latent space.
Specifically, an autoencoder~\cite{kingma2013auto} converts an image $x$ to a latent $z$ with encoder $\mathcal{E}$ and reconstructs it with decoder $\mathcal{D}$, \ie $\hat{x} = \mathcal{D}(\mathcal{E}(x))$.
In diffusion process, Gaussian noise $\epsilon\sim\N(0,1)$ is applied to the starting latent $z_0$ to achieve noisy latent $z_t$:
\begin{equation}
    z_t=\sqrt{\bar{\alpha}_t}z_0+\sqrt{1-\bar{\alpha}_t}\epsilon,
\label{equ:zt}
\end{equation}
where $\{\bar{\alpha}_t\}_{t=1}^T$ is a pre-defined noise variance schedule over timestep $t$.
For denoising, a network $\epsilon_{\theta}$ is optimized to predict the noise $\epsilon$ at a particular timestep $t$ conditioned on $c$, such as text prompt, which follows the objective: 
\begin{equation}
    \mathcal{L}_{ldm} = \mathbb{E}_{z,\epsilon\sim\N(0,1), t} \Big[\Vert \epsilon - \epsilon_{\theta}(z_t,t,c)\Vert^2_2 \Big].
\end{equation}
Given the noise predicted at timestep $t$, the denoised latent $\hat{z}_0$ can be predicted by: 
\begin{equation}
\hat{z}_0 = \frac{1}{\sqrt{\bar{\alpha}_t}} \left( z_t - {\sqrt{1 - \bar{\alpha}_t}} \epsilon_{\theta}(z_t,t,c) \right).
\label{equ:z0}
\end{equation}
The denoised image can be further obtained by $\hat{x}_0 = \mathcal{D}(\hat{z}_0)$.

\noindent{\bf Low-Rank Adaptation (LoRA)}
is widely used for the efficient adaptation of large language models or text-to-image models to downstream tasks~\cite{hu2022lora}.
Concretely, two low-rank matrices $B\in \mathbb{R}^{n\times r}$ and $A\in \mathbb{R}^{r\times m}$ are introduced to update the base model weights $W_0\in \mathbb{R}^{n\times m}$ with $\triangle{W}=BA$, where $r \ll min(n, m)$ refers to the intrinsic rank of $\triangle{W}$.
The updated weight matrix $W=W_0 + \triangle{W}$ is used for inference.
By updating $A$ and $B$ exclusively during training, the base model gradually adapts to the specific domains.
Meanwhile, since $W_0$ is frozen throughout the optimization, the generative ability of base model is well preserved.
For efficiency, LoRA is typically applied only to attention layers for model fine-tuning.

\section{Methodology}
This work aims to handle inherent gaps between the vanilla diffusion prior and blind face restoration (BFR) setting, which stem from the distribution mismatch between 1) high-quality (HQ) and low-quality (LQ) images, and 2) synthesized and real-world images.
Unlike previous attempts addressing the first gap with pre-processing modules~\cite{lin2023diffbir, oh2025pmrf} or guidance from other face restoration methods~\cite{yang2023pgdiff}, our model, namely \ours, is simply built upon a T2I model with few trainable parameters.
The framework is illustrated in~\cref{fig:frame}, which takes pairs of HQ and LQ images as input.
By simply flipping the input, the model switches between \RM (\cref{sec:rm}) and \DM (\cref{sec:dm}) to achieve authentic and faithful face restoration and synthesizing real-world degraded images, respectively.
Note that, to preserve the strong image prior of the large-scale T2I model, we fine-tune only the LoRA weights~\cite{hu2022lora} plugged to all attention layers.

\subsection{Restoration mode}\label{sec:rm}
In \RM, the denoising network takes HQ images $\bm{x}$ as input and LQ images $\bm{y}$ as condition.
Unlike previous works concatenating LQ images~\cite{miao2024waveface} or pre-processed results~\cite{lin2023diffbir,oh2025pmrf} to HQ images, our model is conditioned on diffused LQ images.
Specifically, the diffusion process (\cref{equ:zt}) is applied to both HQ and LQ images, where noise ($\epsilon_x$, $\epsilon_y$) and timestep ($t_x$, $t_y$) are sampled separately.
The noisy pairs $(\bm{x}_{t_x}, \bm{y}_{t_y})$ are concatenated along the batch axis and fed to the denoising network, which is optimized to simultaneously denoise both inputs, enabling the restoration of HQ images from LQ counterparts under varying noise levels.
However, LQ images act as a double-edged sword, which improves the restoration fidelity by providing structural information while compromising the authenticity.
To strike a balance, \jattn is proposed to selectively integrate BFR-oriented features, and meanwhile, discard irrelevant ones.
Additionally, we identify the misalignment between face embeddings widely-used for personalized generation~\cite{papantoniou2024arc2face, peng2024portraitbooth} and those required by BFR, which inspires us to explore BFR-oriented face embeddings. 

\noindent\textbf{\jattn.}
Following the pioneering T2I model~\cite{rombach2022ldm}, our denoising network $\epsilon_{\theta}$ is implemented as UNet, with its blocks comprising ResNet layers, self-attention layers, and cross-attention layers.
Inspired by a joint module used in conditioned generation~\cite{li2024unicon}, we use \jattn to enhance noisy HQ features with BFR-oriented features from LQ conditions.
As shown in~\cref{fig:frame}, \jattn is placed in parallel to self-attention layers to involve LQ features without interrupting the spatial relationships within HQ images.

As illustrated in~\cref{fig:frame}, given the noisy HQ and LQ features ($F_x$, $F_y$), \jattn projects them to query, key and value via projection matrices $W_{Q_x}$, $W_{K_y}$ and $W_{V_y}$ and then integrate them with attention mechanism:
\begin{equation}
    \begin{aligned}
    F_{ro} &= W_O \cdot \text{Attn}(Q_x,K_y,V_y) \\
    &= W_O \cdot \text{Softmax}(Q_xK_{y}^T / \sqrt{d})V_y,
    \end{aligned}
\end{equation}
where $Q_x=W_{Q_x}F_x$, $K_y=W_{K_y}F_y$, and $V_y=W_{V_y}F_y$.
Following prior arts~\cite{guo2023animatediff}, $W_{Q_x}$, $W_{K_y}$ and $W_{V_y}$ are initialized by corresponding self-attention weight of the base model and the output projection matrices $W_O$ are initialized by all-zero matrix to avoid harmful effects that additional modules might introduce.
The BFR-oriented features $F_{ro}$ are then combined with self-attention output $F_{sa_x}$ as:
\begin{equation}
    F_{joint} = F_{sa_x} + \phi \cdot F_{ro} =\text{Self-Attn}(F_x) + \phi \cdot F_{ro},
\label{equ:fjoint}
\end{equation}
where the enhance weight $\phi$ is set to control how much LQ features will be involved.
Subsequently, the joint features $F_{joint}$ are integrated with text prompts as well as BFR-oriented face embeddings via cross-attention.

\noindent\textbf{BFR-oriented face embeddings.}
In personalized generation works~\cite{papantoniou2024arc2face, peng2024portraitbooth}, face embeddings, acting as an additional condition, are integrated via cross-attention layers to control the identity.
However, prior arts generally leverage ID embeddings from a face recognition network~\cite{deng2019arcface}.
Yet, such embeddings are trained to distinguish different persons instead of preserving fine-grained facial details, making them ineffective in providing instructive guidance for restoration.
To this end, this work aims to explore BFR-oriented face embeddings, which should 1) well represent the appearance of the given person, and 2) be positioned in the common latent space of both HQ and LQ inputs so that they will not be biased when integrating with joint features from \jattn in cross-attention layers.
Inspired by the training scheme used in DAEFR~\cite{tsai2024daefr}, we propose to obtain such embeddings from an autoencoder~\cite{kingma2013auto} trained for face reconstruction.
Instead of the discrete codebook, we adopt continuous latent features which can provide fine-grained details.
A two-stage learning scheme is adopted.

\noindent\textbf{a) Reconstruction stage.}
First, HQ/LQ autoencoders are trained to reconstruct input face images.
HQ/LQ images $\bm{x}/\bm{y}\in \mathbb{R}^{H\times W \times 3}$ are projected to latent space $\bm{z}_{x/y} \in \mathbb{R}^{h\times w \times d} $ via encoder and then projected back to images $\hat{\bm{x}}/\hat{\bm{y}}$ by decoder.
To achieve faithful reconstruction, the objective consists of L1 loss $\mathcal{L}_1$, perceptual loss $\mathcal{L}_{p}$, and adversarial loss $\mathcal{L}_{adv}$, \ie
$\mathcal{L}_{ae} = \mathcal{L}_1 + \lambda_{ap} \cdot \mathcal{L}_{p} + \lambda_{adv} \cdot \mathcal{L}_{adv}$,
where $\lambda_{ap}$ and $\lambda_{adv}$ are set as 0.5 and 0.8, respectively.

\noindent\textbf{b) Association stage.}
Similar to CLIP~\cite{radford2021clip}, we fine-tune the autoencoders via the cross-entropy loss to map HQ/LQ features into a common latent space.
HQ and LQ latents are first flattened to $\{ \bm{z}_{x/y}^i \in \mathbb{R}^{d}\}_{i=1}^{N}, N=h\times w$.
The similarity matrix between $\bm{z}_{x}$ and $\bm{z}_{y}$ can be computed as $\mathcal{M}\in \mathbb{R}^{N \times N}$.
We adopt the cross-entropy loss ($\mathcal{L}^{H}_{ce}$ and $\mathcal{L}^{L}_{ce}$) to maximize the similarity between corresponding patches via maximizing scores in the diagonal:
\begin{equation}
\mathcal{L}^{H(L)}_\text{ce} = -\frac{1}{N} \sum_{i=1}^{N} \sum_{j=1}^{N} y_{i,j} \log(z_{x(y)}^{i,j}),
\end{equation}
where $y_{i,j}$ represents ground-truth labels.
The objective of the association stage is $\mathcal{L}_{asso} = \mathcal{L}_{ae}+(\mathcal{L}^{H}_{ce}+\mathcal{L}^{L}_{ce})/2$.

Once trained, the LQ encoder is used to extract face embeddings $\bm{z}_{y} \in \mathbb{R}^{N \times d}$, which are then passed through a lightweight adapter $\tau(\cdot)$ to align with the dimensionality of text tokens before integrating with the T2I model, \ie $c_{\text{id}} = \tau(\bm{z}_y)$.
The adapter consists of several linear layers with Layer Normalization~\cite{ye2023ipadapter}.

\noindent\textbf{Objectives.}
In \RM, the model learns how to denoise the noisy $(\bm{x}_{t_{x}}, \bm{y}_{t_{y}})$ jointly following the objective: 
\begin{equation}
    \mathcal{L}_{ldm} = \mathbb{E}_{\bm{x}, \bm{y}, \epsilon, t_{x}, t_{y}} \Big[\Vert \epsilon - \epsilon_{\theta}(\bm{x}_{t_{x}}, \bm{y}_{t_{y}}, t_x, t_y, c)\Vert^2 \Big],
    \label{eq:dm_obj}
\end{equation}
where $\epsilon = (\epsilon_x, \epsilon_y)$ and $c=(c_{\text{text}},c_{\text{id}})$.
Beyond latent-level constraints, image-level constraints are applied to the denoised predictions $\hat{\bm{x}}_0$ at each timestep obtained by~\cref{equ:z0}, which include MSE loss $\mathcal{L}_{mse}$ and perceptual loss $\mathcal{L}_p$:
\begin{equation}
    \mathcal{L}_{mse}=\| \bm{x} - \hat{\bm{x}}_0\|_2^2, \quad \mathcal{L}_{p} = \| \phi(\bm{x}) - \phi(\hat{\bm{x}}_0) \|_2^2,
\end{equation}
where $\phi(\cdot)$ represents VGG19~\cite{simonyan2014vgg} backbone.
The overall training objective is $\mathcal{L}_{rm}=\mathcal{L}_{ldm} + \lambda_{mse} \cdot \mathcal{L}_{mse} + \lambda_{p} \cdot \mathcal{L}_{p}$.
$\lambda_{mse}$ and $\lambda_{p}$ are set as 1 and 0.01 throughout the training. 

\subsection{Degradation mode}\label{sec:dm}
By simply flipping the input, \ie LQ images $\bm{y}$ serving as input while HQ images $\bm{x}$ as condition, our \ours switches to \DM.
To learn the real-world degradations, our model needs to be trained with paired HQ and LQ images collected from real-world scenarios.  
In the absence of such face datasets, we gather multiple low-level vision datasets captured in real-world scenarios to construct the training dataset for \DM, including Dense-Haze~\cite{ancuti2019densehaze} for image dehazing, LOL~\cite{wei2018lol} for low light enhancement, SIDD~\cite{abdelhamed2018ssid} for image denoising, and RealBlur~\cite{rim2020realblur} for image deblurring.
To construct the training set, full set of Dense-Haze and LOL are included and 1000 pairs are randomly selected from SIDD and RealBlur-J, obtaining a real-world dataset with around 2500 pairs.
To keep the model's ability in synthesizing the classical degradations~\cref{eq:degradation}, 4000 face images are randomly selected from FFHQ to synthesize LQ counterparts on-the-fly during training.
The training process is the same with \RM, except for the condition and the objective.

\noindent{\textbf{Cross-attention conditions.}}
Being trained on both face and non-face datasets, we do NOT leverage BFR-oriented face embeddings in this mode.
In other words, text prompts are the only condition for cross-attention layers, which are generated by BLIP~\cite{li2022blip} following previous works~\cite{guo2023animatediff}.

\noindent{\textbf{Objectives.}}
Only diffusion loss $\mathcal{L}_{ldm}$ is used, \ie $\mathcal{L}_{dm} = \mathcal{L}_{ldm}$, as imposing image-level constraints on degraded images will introduce unforeseen noise to predictions.

Once trained, the model can be used to synthesize degraded counterparts of HQ face images with the acquired knowledge.
The generated images carry diverse and complex degradations, with the intensity of degradation remaining entirely unknown, which are more suitable for ``BLIND'' face restoration.
The visualization and distribution of degraded images synthesized by a prior art~\cite{wang2021realesrgan} and our method are illustrated in~\cref{fig:tsne}.
As observed, \ours could well model real-world degradations, evidenced by its diverse distribution across multiple real-world face datasets.
The synthesized degraded images can be subsequently used to train our \ours under \RM.

\vspace{-1mm}
\section{Experiments}
\begin{figure*}
    \centering
    \includegraphics[width=\linewidth]{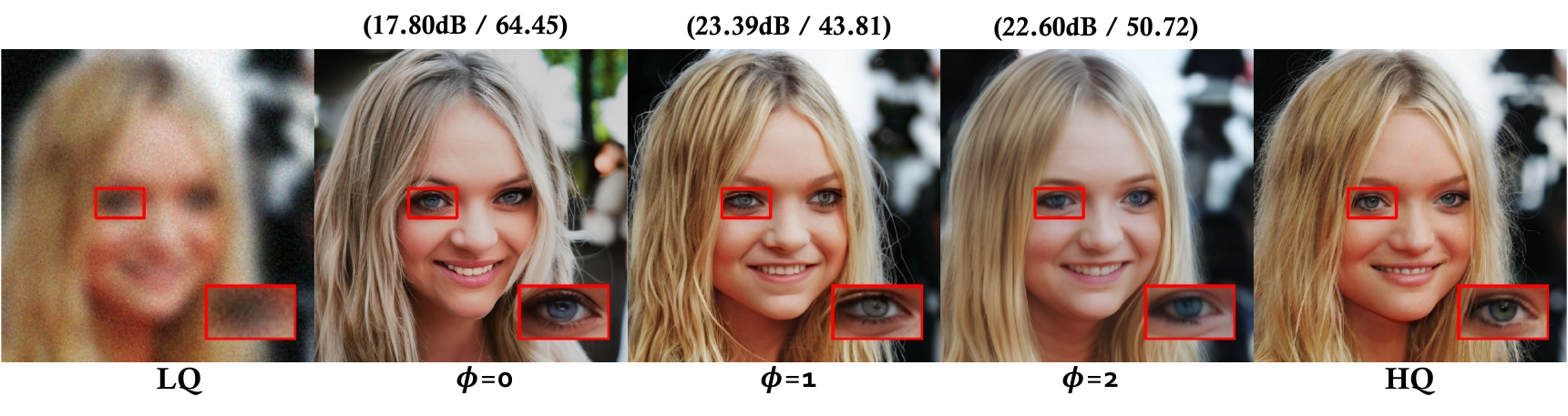}
    \vspace{-8mm}
    \caption{Visualization of the effectiveness of enhance weight $\phi$ on CelebA-Test. $\phi$ strikes a balance between fidelity and authenticity by determining the amount of LQ features involved. PSNR($\uparrow$) / Deg.($\downarrow$) are reported.}
    \label{fig:ab_hubw}
    \vspace{-4mm}
\end{figure*}

\subsection{Datasets and Settings} \label{sec:dataset}
\noindent\textbf{Training Dataset.}
FFHQ~\cite{karras2019stylegan}, containing 70k high-quality (HQ) face images, is used as the training set.
Following previous works~\cite{wang2021gfpgan, wang2023dr2}, images are resized to $512\times 512$ before synthesis.
Two degradation models are adopted to synthesize low-quality (LQ) counterparts:

\noindent\textbf{1) On-the-fly degradation.}
Considering its efficiency, Real-ESRGAN~\cite{wang2021realesrgan} is used to synthesize degraded images online following the formula:
\begin{equation}
    \bm{y} = \left\{\left[\left(\bm{x} \otimes \bm{k}_{\sigma}\right){\downarrow_s} + \bm{n}_{\delta}\right]_{\text{JPEG}_q}\right\}{\uparrow_s},
    \label{eq:degradation}
\end{equation}
where a HQ image $\bm{x}$ is firstly blurred by a Gaussian kernel $\bm{k}_{\sigma}$, followed by a downsampling of scale $s$.
Afterward, Gaussian noise $\bm{n}_{\delta}$ and JPEG compression with quality factor $q$ are applied to the image, which is then upsampled back to the original size to obtain its LQ counterpart $\bm{y}$. 
The hyper-parameters $\sigma$, $s$, $\delta$, and $q$ are uniformly sampled from $[0.1,15]$, $[0.8,32]$, $[0,20]$, and $[30,95]$ respectively.
On top of this, the process is performed for the second time, where random degradations are applied in shuffled order.

\noindent\textbf{2) Off-shelf degradation.}
In light of its diversity and authenticity, our \ours is switched to \DM to synthesize degraded images offline.
For each HQ face image, we generate five images with diverse and unknown degradations, serving as LQ counterparts.  

For each iteration, degraded images generated by on-the-fly and off-shelf schemes are selected by a probability of 0.5 to construct a training batch.

\noindent\textbf{Testing Dataset.}
We evaluate \ours on a synthetic dataset: CelebA-Test and three real-world datasets: LFW~\cite{huang2008labeled}, WebPhoto~\cite{wang2021gfpgan} and WIDER~\cite{zhou2022codeformer}.
CelebA-Test contains 3000 HQ images from CelebA-HQ~\cite{karras2018progressive}.
Akin to the training set, LQ counterparts in test set are synthesized by both on-the-fly and off-shelf degradation scheme at equal probability.
In terms of three real-world datasets,
LFW contains 1711 mildly degraded face images in the wild, which comprises the first image for each person in LFW~\cite{huang2008labeled}.
WebPhoto includes 407 images crawled from the internet, some of which are old photos with severe degradation.
WIDER consists of 970 images with severe degradations from the WIDER dataset~\cite{yang2016wider}.

\noindent\textbf{Evaluation Metrics.}
For evaluation, we adopt two pixel-wise metrics (PSNR and SSIM), a reference perceptual metric (LPIPS~\cite{zhang2018lpips}), and a non-reference perceptual metric (FID~\cite{heusel2017fid}). 
To measure the consistency of identity, the angle between ID embeddings extracted by ArcFace (``Deg.'')~\cite{deng2019arcface} is used.
All metrics are used for the evaluation of synthetic dataset while only FID is used for real-world datasets due to the lack of ground-truth HQ images.

\noindent\textbf{Implementation Details.}
We adopt Stable Diffusion 2.1-base\footnote{\scriptsize Stable Diffusion v2.1: \url{https://github.com/Stability-AI/stablediffusion}} as the base T2I model and plug LoRA modules to all self-attention and cross-attention layers. 
We train LoRA for 90k iterations (batch size=16), with the intrinsic rank set to 64.
The text prompts are generated by BLIP~\cite{li2022blip} for all datasets.
We use AdamW~\cite{kingma2015adam} optimizer with the learning rate 5e-5.
The training process is conducted on $512\times512$ resolution with 8 NVIDIA V100 GPUs.

\noindent For \RM, autoencoders are initialized by the pretrained weight~\cite{rombach2022ldm}.
We trained model for 200k at reconstruction stage and 80k at association stage (batch size=32).
For non-face datasets, the model takes no embeddings other than text prompts as condition.
During training, we use a probability of 0.5 to drop text prompt and 0.3 to drop face embeddings to enhance the robustness.
For inference, we use words such as ``low quality'', ``blurry'', and ``poorly rendered hands'' as the negative prompt for \RM while use null text prompt for \DM.
The classifier-free guidance (CFG) scale is set as 1.5.

\subsection{Ablation Studies}\label{sec:ab_study}
\subsubsection{Restoration mode} \label{sec:ab_rm}
\noindent \textbf{\jattn enhance weight.}
According to~\cref{equ:fjoint}, the enhance weight $\phi$ determines the amount of BFR-oriented features integrated from LQ inputs.
When less LQ features are involved, the ability of the T2I model in generating high-quality contents is fully exploited while features that facilitate face restoration, such as color and structure, are ignored, thereby affecting the authenticity.
Conversely, an over-injection of such features will compromise the restoration quality.
Restoration results with different enhance weights as illustrated in~\cref{fig:ab_hubw}.

As observed, without guidance from LQ images ($\phi=0$), the restoration strictly relies on text prompt and face embeddings.
Despite its high quality, the restoration deviates significantly from the original identity.
In contrast, LQ features dominate the generation when $\phi=2$, leading to the poor quality with fine-grained facial details being smoothed out.
When we set $\phi=1$, the model could strike a good balance between fidelity and authenticity.

\begin{figure*}
    \centering
    \includegraphics[width=\linewidth]{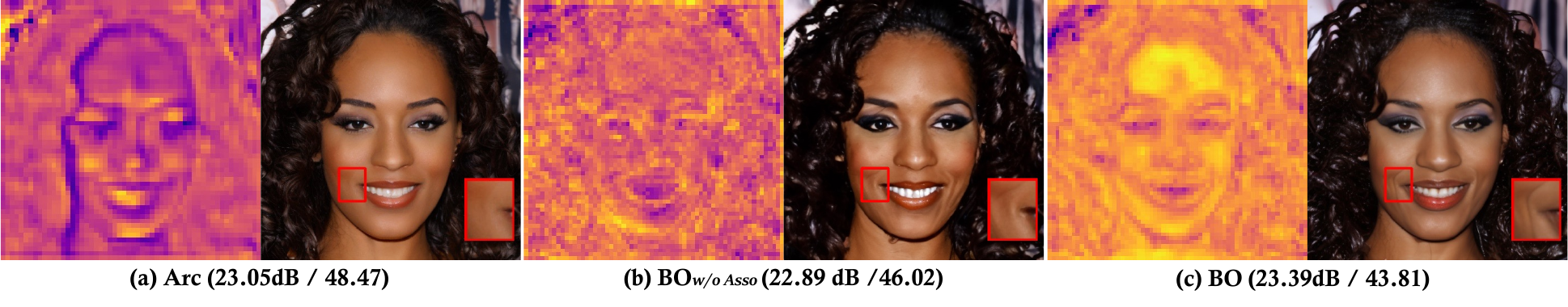}
    \vspace{-8mm}
    \caption{Restoration results (Right) of the model conditioning on face embeddings extracted from ArcFace (Arc), our BFR-oriented ones trained without (BO$_{w/o\,Asso}$) and with (BO) ``Association stage'' on CelebA-Test. Corresponding attention maps (Left) of embeddings are also shown. PSNR($\uparrow$) / Deg.($\downarrow$) are reported. The lighter the color, the higher the attention scores.}
    \label{fig:ab_idemb}
    \vspace{-4mm}
\end{figure*}

\noindent \textbf{BFR-oriented face embeddings.}
We identify the misalignment between the widely-used face embeddings from a face recognition model~\cite{deng2019arcface} and those required by blind face restoration (BFR).
To make a comparison, we visualize the restoration results, conditioned on ArcFace embeddings and our BFR-oriented face embeddings in~\cref{fig:ab_idemb}(a) and~\cref{fig:ab_idemb}(c), as well as corresponding attention maps.
As observed, the face recognition model, trained to distinguish different persons, exclusively focuses on features at discriminative regions, such as eyes.
As a result, the restoration fails to render fine-grained details at cheeks or hairs.

Beyond presenting facial details, the alignment between HQ and LQ face embeddings is important for our model as face embeddings are supposed to integrated with joint features (\cref{equ:fjoint}) via cross-attention, which include both HQ features from self-attention layers and BFR-oriented LQ features from \jattn.
The alignment is achieved by ``Association stage''.
To demonstrate its effectiveness, we use face embeddings from the autoencoder trained only with reconstruction stage to perform the restoration. The result is shown in~\cref{fig:ab_idemb}(b).
Due to the absence of ``Association stage'', face embeddings are biased on BFR-oriented LQ features while neglecting HQ ones, leading to compromised the restoration quality.
On the contrary, when BFR-oriented face embeddings integrating with the aligned HQ and LQ features, our model is able to deliver faithful results with face details such as wrinkle well preserved (\cref{fig:ab_idemb}(c)).

\begin{figure}
    \centering
    \includegraphics[width=\linewidth]{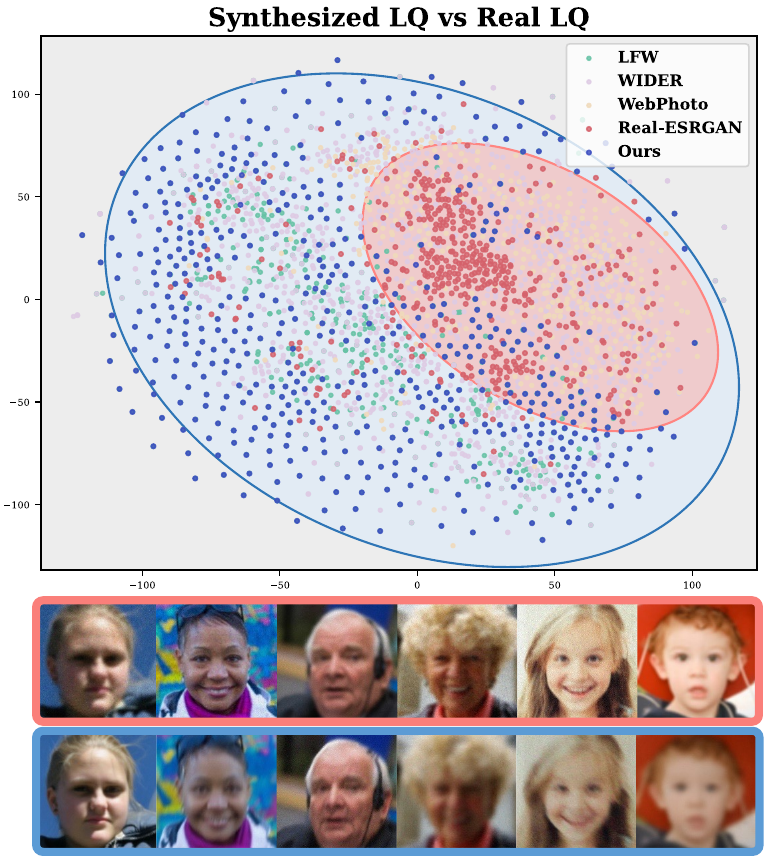}
    \vspace{-6mm}
    \caption{Distribution of degraded images synthesized by Real-ESRGAN~\cite{wang2021realesrgan} and our \ours compared to three real-world datasets: LFW, WebPhoto and WIDER. Synthesized images of \textcolor{realpink}{\textbf{Real-ESRGAN}} and \textcolor{ourblue}{\textbf{Ours}} are shown below.}
    \label{fig:tsne}
    \vspace{-6mm}
\end{figure}

\subsubsection{Degradation mode} \label{sec:ab_dm}
\noindent\textbf{Synthesized images and distribution.}
We visualize the degraded images generated by Real-ESRGAN~\cite{wang2021realesrgan} and our \ours in~\cref{fig:tsne}.
To evaluate how well they fit real-world degradations, the feature distribution of degraded images synthesized by both methods and three real-world datasets: LFW~\cite{huang2008labeled}, WIDER~\cite{zhou2022codeformer} and WebPhoto~\cite{wang2021gfpgan} is also illustrated.
Our model is capable of synthesizing degraded images that are widely distributed across different real-world datasets whereas Real-ESRGAN generates data gathering in a limited region.
This can be proved by examples shown below.
Images degraded by our model present diverse degradations while that by Real-ESRGAN demonstrate limited degradation patterns.

\begin{table}[h]
\vspace{-1mm}
\small
\centering
\caption{Quantitative comparison on \textbf{CelebA-Test}. ``RF++'' and ``$\mathcal{F}$'' refer to ``RestoreFormer++'' and our \ours. \onres{``$+$O''}: on-the-fly synthesized images. \onour{``$+$O/F''}: both on-the-fly and off-shelf synthesized images. ``Deg.'' refers to the angle between identity embeddings of restored images and HQ counterparts. The \bred{best} and the \underline{second best} results are indicated.}
\tabcolsep=0.1cm
\vspace{-1mm}
\scalebox{0.84}{
\begin{tabular}{c|ccccc}
    \toprule
    \textbf{Methods} & \textbf{PSNR}$\uparrow$ & \textbf{SSIM}$\uparrow$ & \textbf{LPIPS}$\downarrow$  & \textbf{FID}$\downarrow$ & \textbf{Deg.}$\downarrow$    \\ 
    \midrule
    GPEN & \onres{23.65}/\onour{22.86} & \onres{0.61}/\onour{0.60} & \onres{0.42}/\onour{0.43} & \onres{18.67}/\onour{20.20} & \onres{47.24}/\onour{51.33} \\
    GFP-GAN & \onres{23.85}/\onour{22.70} & \onres{0.61}/\onour{0.61} & \onres{0.40}/\onour{0.40} & \onres{17.38}/\onour{18.52} & \onres{44.96}/\onour{47.99} \\
    \hdashline
    RF++ & \onres{24.03}/\onour{23.15} & \onres{0.63}/\onour{0.62} & \onres{\underline{0.38}}/\onour{\underline{0.40}} & \onres{14.98}/\onour{\underline{15.15}} & \onres{44.00}/\onour{45.86} \\
    DAEFR & \onres{22.70}/\onour{21.96} &\onres{0.62}/\onour{0.61} & \onres{\bred{0.38}}/\onour{\bred{0.39}} & \onres{14.43}/\onour{15.65} & \onres{48.44}/\onour{49.85} \\
    \hdashline
    DR2 & \onres{22.27}/\onour{21.83} & \onres{0.63}/\onour{0.61} & \onres{0.44}/\onour{0.44} & \onres{31.59}/\onour{34.49} & \onres{59.24}/\onour{61.29} \\
    PGDiff & \onres{22.87}/\onour{22.08} & \onres{0.63}/\onour{0.62} & \onres{0.42}/\onour{0.42} & \onres{19.48}/\onour{21.35} & \onres{59.43}/\onour{60.38} \\
    DiffBIR & \onres{24.37}/\onour{23.39} & \onres{0.62}/\onour{0.60} & \onres{0.42}/\onour{0.43} & \onres{20.78}/\onour{22.54} & \onres{46.58}/\onour{47.00} \\
    WaveFace & \onres{24.16}/\onour{23.05} &\onres{0.62}/\onour{0.62} & \onres{0.42}/\onour{0.43} & \onres{21.44}/\onour{23.48} & \onres{\underline{43.80}}/\onour{46.50} \\ 
    PMRF & \onres{\bred{24.70}}/\onour{\bred{23.40}} & \onres{\underline{0.63}}/\onour{\underline{0.62}} & \onres{0.42}/\onour{0.42}& \onres{\bred{13.48}}/\onour{15.36} & \onres{44.75}/\onour{\underline{45.76}} \\
    \midrule
    \bgcgrey{\textbf{$\mathcal{F}$}\onres{\tiny{$+$O}}\scriptsize{/}\onour{\tiny{$+$O/F}}} & \bgcgrey{\onres{\underline{24.44}}/\onour{\underline{23.39}}} & \bgcgrey{\onres{\bred{0.64}}/\onour{\bred{0.63}}} &\bgcgrey{\onres{0.42}/\onour{0.42}} & \bgcgrey{\onres{\underline{13.82}}/\onour{\bred{15.06}}} & \bgcgrey{\onres{\bred{43.29}}/\onour{\bred{43.81}}} \\ 
    \bottomrule
\end{tabular}}
\vspace{-0.2cm}
 \label{tab:celeba_blind}
\end{table}

\begin{figure*}[!htbp]
\begin{subfigure}{0.123\textwidth}
\includegraphics[height=6cm]{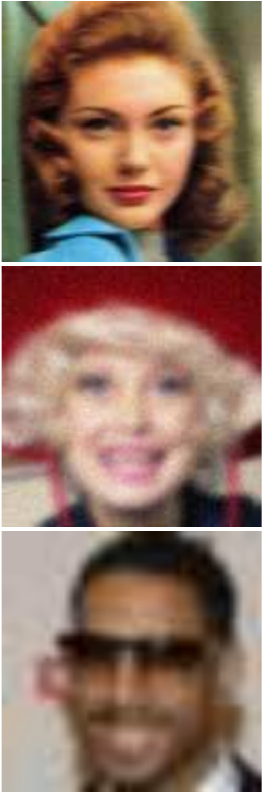} 
\subcaption*{\textbf{LQ}}
\end{subfigure}
\hspace{-4mm}
\begin{subfigure}{0.123\textwidth}
\includegraphics[height=6cm]{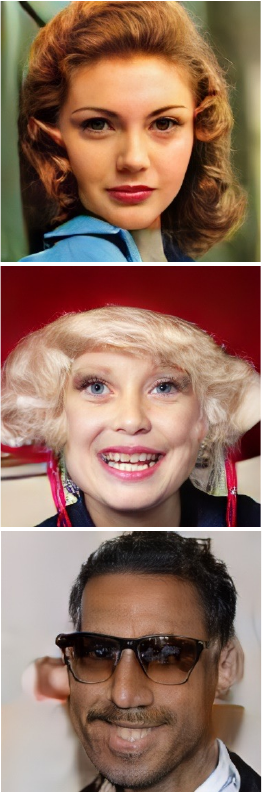} 
\subcaption*{GFPGAN~\cite{wang2021gfpgan}}
\end{subfigure}
\hspace{-4mm}
\begin{subfigure}{0.123\textwidth}
\includegraphics[height=6cm]{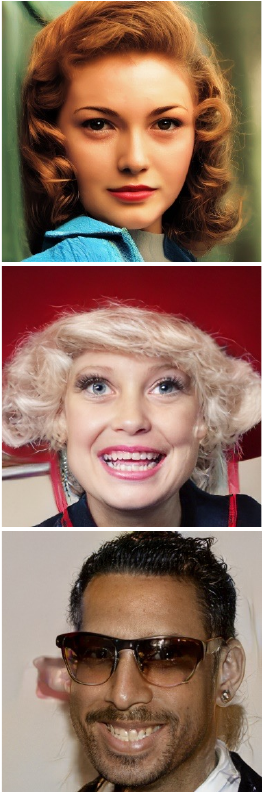} 
\subcaption*{RF++~\cite{wang2023restoreformer++}}
\end{subfigure}
\hspace{-4mm}
\begin{subfigure}{0.123\textwidth}
\includegraphics[height=6cm]{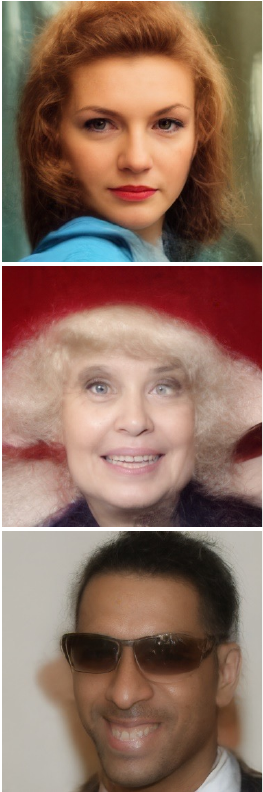} 
\subcaption*{DR2~\cite{wang2023dr2}}
\end{subfigure}
\hspace{-4mm}
\begin{subfigure}{0.123\textwidth}
\includegraphics[height=6cm]{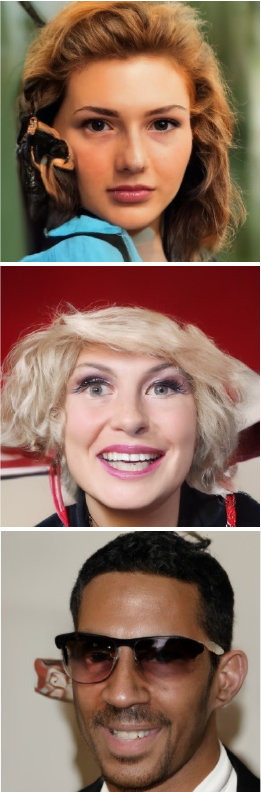} 
\subcaption*{PGDiff~\cite{yang2023pgdiff}}
\end{subfigure}
\hspace{-4mm}
\begin{subfigure}{0.123\textwidth}
\includegraphics[height=6cm]{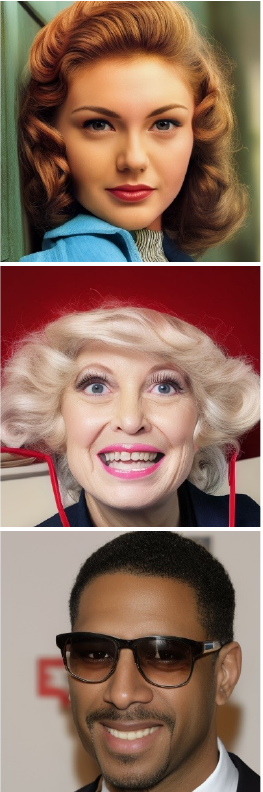} 
\subcaption*{DiffBIR~\cite{lin2023diffbir}}
\end{subfigure}
\hspace{-4mm}
\begin{subfigure}{0.123\textwidth}
\includegraphics[height=6cm]{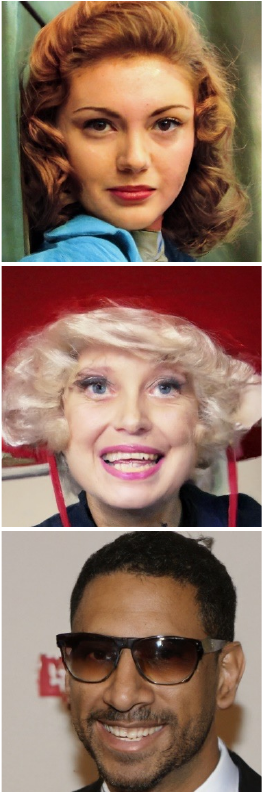} 
\subcaption*{PMRF~\cite{oh2025pmrf}}
\end{subfigure}
\hspace{-4mm}
\begin{subfigure}{0.123\textwidth}
\includegraphics[height=6cm]{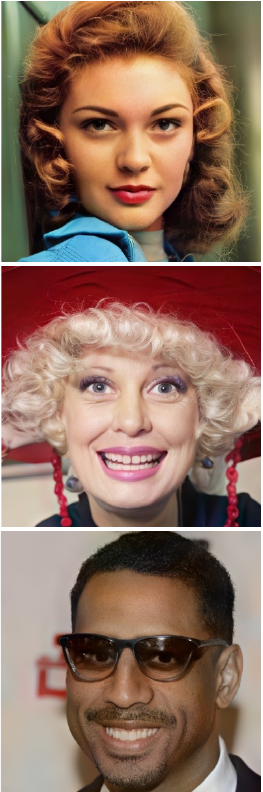} 
\subcaption*{\textbf{Ours}}
\end{subfigure}
\hspace{-4mm}
\begin{subfigure}{0.123\textwidth}
\includegraphics[height=6cm]{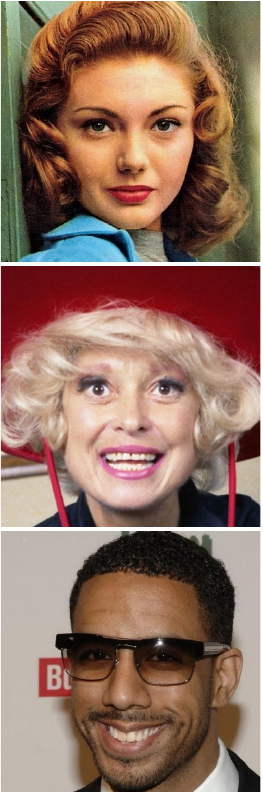} 
\subcaption*{\textbf{HQ}}
\end{subfigure}
\hspace{-4mm}
\vspace{-2mm}
\caption{Qualitative comparison with state-of-the-art BFR methods on \textbf{CelebA-Test}. Our method achieves authentic and faithful restoration with fine-grained facial details being well preserved. (Zoom in for best view).}
\label{fig:syn}
\vspace{-4mm}
\end{figure*}
\begin{figure*}[!htbp]
\begin{subfigure}{0.123\textwidth}
\includegraphics[height=6cm]{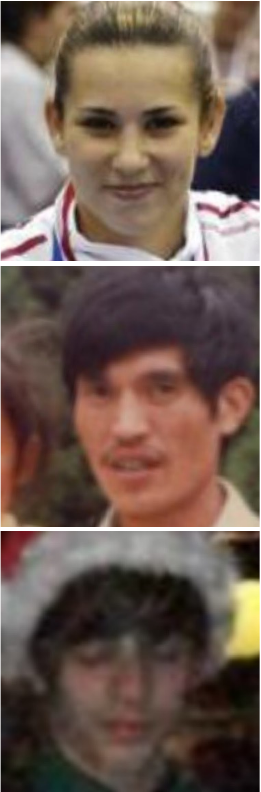} 
\subcaption*{\textbf{LQ}}
\end{subfigure}
\hspace{-4mm}
\begin{subfigure}{0.123\textwidth}
\includegraphics[height=6cm]{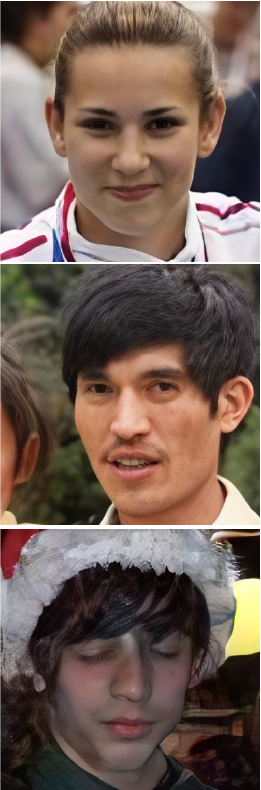} 
\subcaption*{GFPGAN~\cite{wang2021gfpgan}}
\end{subfigure}
\hspace{-4mm}
\begin{subfigure}{0.123\textwidth}
\includegraphics[height=6cm]{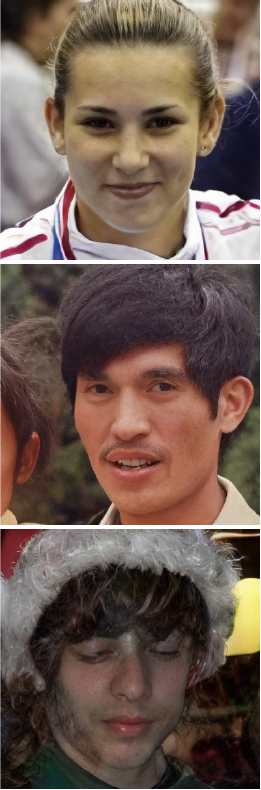} 
\subcaption*{RF++~\cite{wang2023restoreformer++}}
\end{subfigure}
\hspace{-4mm}
\begin{subfigure}{0.124\textwidth}
\includegraphics[height=6cm]{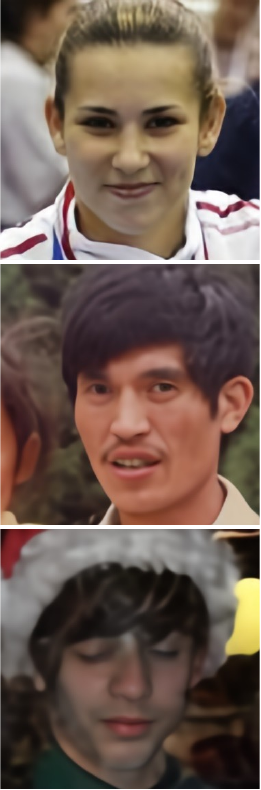} 
\subcaption*{SwinIR~\cite{liang2021swinir}}
\end{subfigure}
\hspace{-4mm}
\begin{subfigure}{0.123\textwidth}
\includegraphics[height=6cm]{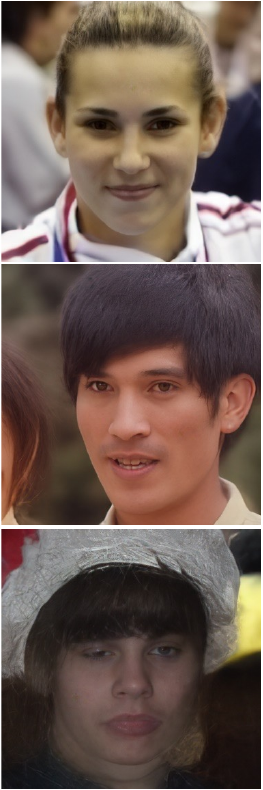} 
\subcaption*{DR2~\cite{wang2023dr2}}
\end{subfigure}
\hspace{-4mm}
\begin{subfigure}{0.123\textwidth}
\includegraphics[height=6cm]{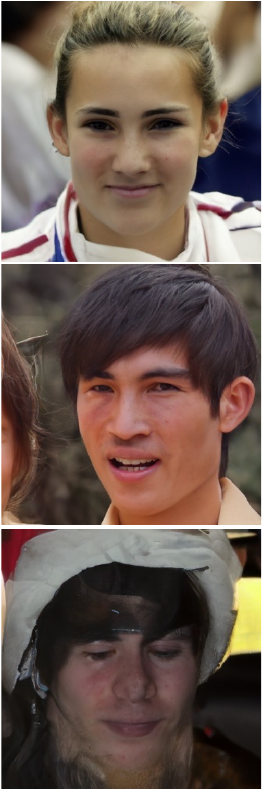} 
\subcaption*{PGDiff~\cite{yang2023pgdiff}}
\end{subfigure}
\hspace{-4mm}
\begin{subfigure}{0.123\textwidth}
\includegraphics[height=6cm]{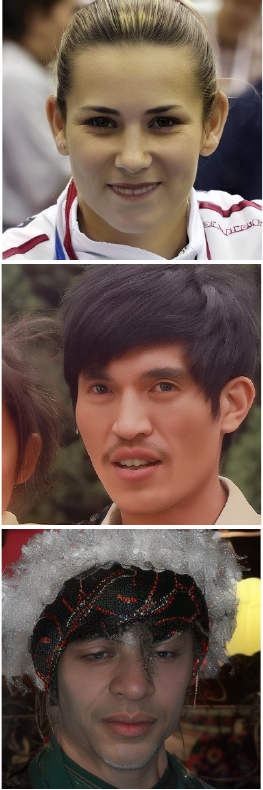} 
\subcaption*{DiffBIR~\cite{lin2023diffbir}}
\end{subfigure}
\hspace{-4mm}
\begin{subfigure}{0.124\textwidth}
\includegraphics[height=6cm]{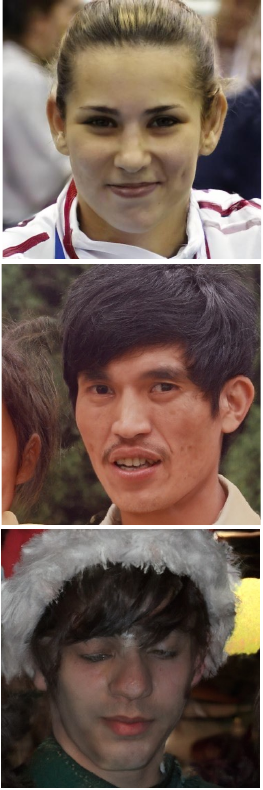} 
\subcaption*{PMRF~\cite{oh2025pmrf}}
\end{subfigure}
\hspace{-4mm}
\begin{subfigure}{0.123\textwidth}
\includegraphics[height=6cm]{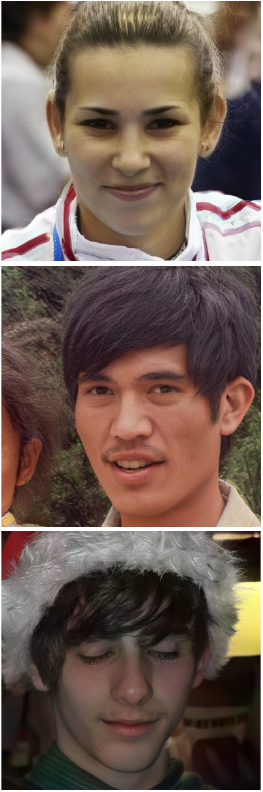} 
\subcaption*{\textbf{Ours}}
\end{subfigure}
\hspace{-4mm}
\vspace{-2mm}
\caption{Qualitative comparison with state-of-the-art BFR methods on \textbf{real-world} datasets, including LFW (first row), WebPhoto (second row), and WIDER (third row). (Zoom in for best view).}
\vspace{-5mm}
\label{fig:real}
\end{figure*}
\begin{table}
\vspace{-1mm}
  \small
  \centering
    \caption{Quantitative comparisons on \textbf{real-world datasets} in terms of \textbf{FID}($\downarrow$). ``RF++'' and ``$\mathcal{F}$'' refer to ``RestoreFormer++'' and our \ours. \onres{``$+$O''}: on-the-fly synthesized images. \onour{``$+$O/F''}: both on-the-fly and off-shelf synthesized images. The \bred{best} and the \underline{second best} results are indicated.}
    \vspace{-1mm}
    \scalebox{0.9}{
    \begin{tabular}{c|c|c|c}
    \toprule
    \textbf{Methods} & \textbf{LFW} & \textbf{WebPhoto} & \textbf{WIDER} \\    
    \midrule 
    \bgcgrey{Input} & \bgcgrey{124.974} & \bgcgrey{170.112} & \bgcgrey{199.961} \\
    \midrule
    GPEN~\cite{yang2021gpen} & 50.792 & 80.572 & 46.340 \\
    GFP-GAN~\cite{wang2021gfpgan} & 49.560 & 87.584  & 39.499 \\
    \hdashline
    RF++~\cite{wang2023restoreformer++} & 50.439 & \bred{75.059} & 49.395 \\
    DAEFR~\cite{tsai2024daefr} & 47.310 & \underline{75.453} & 35.344 \\
    \hdashline
    DR2~\cite{wang2023dr2} & 45.298 &  112.344 & 45.348  \\
    PGDiff~\cite{yang2023pgdiff} & 44.630 & 89.754 & 36.807 \\
    DiffBIR~\cite{lin2023diffbir} & 44.383 & 91.777 & 35.343 \\
    WaveFace~\cite{miao2024waveface} & 43.175 & 81.525 & 36.913 \\
    PMRF~\cite{oh2025pmrf} & 50.275 & 81.064 & 40.685\\
    \midrule
    \bgcgrey{\textbf{$\mathcal{F}$}\onres{\tiny{$+$O}}\scriptsize{/}\onour{\tiny{$+$O/F}}} & \bgcgrey{\onres{\underline{43.10}}/\onour{\bred{42.98}}} & \bgcgrey{\onres{83.34}/\onour{81.37}} & \bgcgrey{\onres{\underline{33.09}}/\onour{\bred{31.81}}} \\ 
    \bottomrule
  \end{tabular}}
  \vspace{-5mm}
  \label{tab:real_fid}
\end{table}
\vspace{-1mm}
\subsection{Comparisons with State-of-the-Art Methods}\label{sec:sota}
\vspace{-1mm}
Evaluations are performed on synthetic and real-world datasets.
Unlike previous BFR methods, our training data are synthesized by dual degradation schemes (on-the-fly and off-shelf).
To make a fair comparison, we also synthesize another group with only on-the-fly degradation scheme.
The quantitative results are indicated by \onour{``$+$O/F''} and \onres{``$+$O''} in~\cref{tab:celeba_blind} and~\cref{tab:real_fid}. 

\noindent\textbf{Synthetic dataset.}
Quantitative comparisons on CelebA-Test~\cite{karras2018progressive} are illustrated in~\cref{tab:celeba_blind}.
Methods tend to yield inferior performance on the test set synthesized by dual degradation schemes, as the degradation better aligns with real-world scenarios, in other words, are more challenging.

For visualization, we present qualitative comparisons with SOTA methods with increasing degradations in~\cref{fig:syn}.
GFPGAN~\cite{wang2021gfpgan} and RestoreFormer++~\cite{wang2023restoreformer++} struggle with obvious artifacts especially when images are corrupted by mild or severe degradation.
They also fail to preserve the identity as the restoration is based on, and also, limited by the StyleGAN prior~\cite{wang2021gfpgan} or the pre-constructed dictionary~\cite{wang2023restoreformer++}.
As inherent gaps between the vanilla diffusion models and the BFR setting have not been properly handled, previous diffusion prior based methods generally yield unsatisfactory results.
DR2~\cite{wang2023dr2} and PGDiff~\cite{yang2023pgdiff} tend to deliver either faithful or authentic results, as the restoration starts from a preliminary result from another face restoration model, making their method suboptimal.
DiffBIR~\cite{lin2023diffbir} and PMRF~\cite{oh2025pmrf}, based on a pre-processing module, typically suffer from rendering vivid facial details or avoiding unexpected artifacts.
In comparison, our model delivers authentic and faithful results.

\noindent\textbf{Real-world datasets.} \label{sec:real_fid}
We show the quantitative comparison on real-world datasets in~\cref{tab:real_fid}.
\ours outperforms SOTA methods on LFW~\cite{huang2008labeled} and WIDER~\cite{zhou2022codeformer}.
Qualitative comparisons (\cref{fig:real}) show that previous diffusion prior based methods yield over-smoothed faces and fail to preserve discriminative facial details.
For severely-degraded images (Row 3), most methods are plagued by artifacts caused by unknown and complex real-world degradation.
In contrast, our approach delivers faithful and authentic restoration results with rich facial details such as mustache.
\vspace{-3mm}
\vspace{-2.5mm}
\section{Conclusion}
We propose \ours to bridge the inherent gap between vanilla diffusion model and BFR settings.
The model provides a two-pronged solution, enabling authentic and faithful face restoration and real-world degradations modeling by simply switching between restoration mode and degradation mode.

{
    \small
    \bibliographystyle{ieeenat_fullname}
    \bibliography{main}
}

\end{document}